\documentclass[runningheads]{llncs}
\usepackage{graphicx}

\usepackage{tikz}
\usepackage{comment}
\usepackage{amsmath,amssymb} 
\usepackage{color}

\usepackage{graphicx}
\usepackage{amsmath}
\usepackage{amssymb}
\usepackage{booktabs}
\usepackage{multirow}
\usepackage{makecell}
\usepackage{diagbox}
\usepackage{slashbox}
\usepackage[OT1]{fontenc}
\usepackage{xcolor}
\usepackage[pagebackref,breaklinks,colorlinks,
citecolor=blue,linkcolor=blue,urlcolor=blue]{hyperref}
\usepackage{cite}
\usepackage{colortbl}
\usepackage{paralist}

\def\etc{\emph{etc}}

\newcommand{\PreserveBackslash}[1]{\let\temp=\\#1\let\\=\temp}
\newcolumntype{C}[1]{>{\PreserveBackslash\centering}p{#1}}
\newcolumntype{R}[1]{>{\PreserveBackslash\raggedleft}p{#1}}
\newcolumntype{L}[1]{>{\PreserveBackslash\raggedright}p{#1}}
\DeclareMathOperator*{\argmin}{argmin}
\newcommand{\ceil}[1]{\lceil {#1} \rceil}
\newcommand{\paragraphx}[1]{\vspace{1mm}\noindent\textbf{#1}}

\usepackage[capitalize]{cleveref}
\crefname{section}{Sec.}{Secs.}
\Crefname{section}{Section}{Sections}
\Crefname{table}{Table}{Tables}
\crefname{table}{Tab.}{Tabs.}

\begin{document}
\pagestyle{headings}
\mainmatter
\def\ECCVSubNumber{}  

\title{CA-SSL: Class-Agnostic Semi-Supervised Learning for Detection \& Segmentation}

\titlerunning{Class-Agnostic Semi-Supervised Learning for Detection \& Segmentation}
%
\author{
Lu Qi$^{1,3}$, Jason Kuen$^{2}$, Zhe Lin$^{2}$, Jiuxiang Gu$^{2}$, Fengyun Rao$^{1}$, Dian Li$^{1}$, Weidong Guo$^{1}$\thanks{Weidong Guo is the corresponding author.}, Zhen Wen$^{1}$, Ming-Hsuan  Yang$^{3}$,  Jiaya Jia$^{4}$ \\
$^{1}$QQ Browser Lab, Tencent  \\
$^{2}$Adobe Research \\
$^{3}$The University of California, Merced \\
$^{4}$The Chinese University of Hong Kong
}
\institute{}
\authorrunning{Lu Qi et al.}
\maketitle

\begin{abstract}
To improve instance-level detection/segmentation performance, existing self-supervised and semi-supervised methods extract either task-unrelated or task-specific training signals from unlabeled data. 
We show that these two approaches, at the two extreme ends of the task-specificity spectrum, are suboptimal for the task performance. Utilizing too little task-specific training signals causes underfitting to the ground-truth labels of downstream tasks, while the opposite causes overfitting to the ground-truth labels. To this end, we propose a novel \textbf{C}lass-\textbf{A}gnostic \textbf{S}emi-\textbf{S}upervised \textbf{L}earning (CA-SSL) framework to achieve a more favorable task-specificity balance in extracting training signals from unlabeled data. CA-SSL has three training stages that act on either ground-truth labels (labeled data) or pseudo labels (unlabeled data). 
This decoupling strategy avoids the complicated scheme in traditional SSL methods that balances the contributions from both data types. Especially, we introduce a warmup training stage to achieve a more optimal balance in task specificity by ignoring class information in the pseudo labels, while preserving localization training signals. As a result, our warmup model can better avoid underfitting/overfitting when fine-tuned on the ground-truth labels in detection and segmentation tasks. Using 3.6M unlabeled data, we achieve a significant performance gain of $4.7\%$ over ImageNet-pretrained baseline on FCOS object detection. 
In addition, our warmup model demonstrates excellent transferability to other detection and segmentation frameworks.
\vspace{-2mm}
\end{abstract}


\vspace{-10mm}
\section{Introduction}\label{sec:intro}
\vspace{-2mm}
Deep learning~\cite{krizhevsky2017imagenet,simonyan2014very,szegedy2015going,szegedy2016rethinking,DBLP:conf/cvpr/HeZRS16,szegedy2016inception,huang2017densely,hu2018squeeze} has enabled instance-level detection (object detection~\cite{girshick2014rich,dollar2015fast,DBLP:conf/nips/RenHGS15}, instance segmentation~\cite{he2017mask,DBLP:journals/corr/abs-1803-01534}, \textit{\etc.}) methods to achieve previously unattainable performance. 
Such success cannot be achieved without large datasets with instance annotations such as COCO~\cite{lin2014microsoft}, Cityscapes~\cite{cordts2016cityscapes} and Open Images~\cite{krasin2016openimages}. However, annotating instances is laboriously expensive due to the great intricateness needed for annotating instance-level bounding boxes, masks, and/or semantic classes. Due to such a limitation, the datasets in instance-level detection domain are relatively small in scale, compared to other domains. 
As such, instance-level detection models generally have degraded generalization performance in real-world applications~\cite{morrison2018cartman,shu2014human,qi2019amodal,chu2019vehicle}.

\begin{figure}[t]
  \centering
   \includegraphics[width=.9\linewidth]{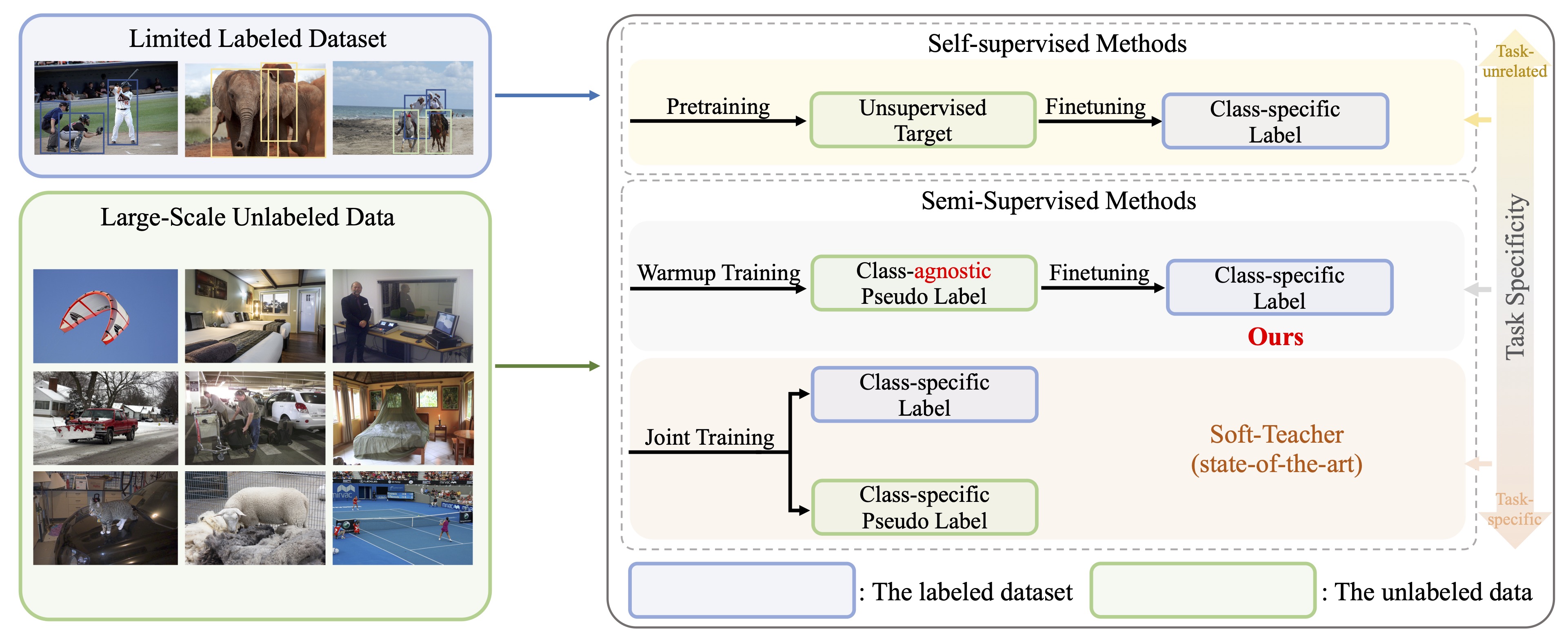}
   \vspace{-4mm}
   \caption{Schematic comparison between existing semi-supervised methods and CA-SSL. Instead of joint training with labeled and unlabeled data, we train on unlabeled and labeled dataset in different stages: \textbf{pseudo labeler training}, \textbf{warmup training}, and \textbf{finetuning}. Furthermore, we use class-agnostic pseudo labels during warmup training.}
   \vspace{-6mm}
   \label{fig:sketch}
\end{figure}

To alleviate the heavy cost of data annotation, numerous methods have been developed to leverage unlabeled images for instance-level detection~\cite{lin2014microsoft,cordts2016cityscapes}. 
Unlike
labeled data that can provide explicit and indisputable supervision signals, the extraction of training signals from unlabeled data remains as an open issue. 
Two widely-adopted approaches for learning from unlabeled data are self-supervised and semi-supervised learning. Self-supervised methods~\cite{he2020momentum,chen2021exploring,xie2021detco,henaff2021efficient,wang2021dense,ramanathan2021predet,dai2021up} usually rely on training signals like the relative distances between the augmented samples of a positive/negative pair, while semi-supervised strategies~\cite{jeong2019consistency,tang2021proposal,radosavovic2018data,zoph2020rethinking,li2020improving,sohn2020simple,wang2018towards,xu2021end} directly leverage the pseudo labels generated by a detector pretrained on ground-truth labels. In terms of task specificity, these two approaches are at the two extreme ends of the spectrum -- self-supervised learning utilizes hardly any task-specific training signals from unlabeled data, whilst semi-supervised learning utilizes too much of them. Consequently, the model tends to underfit/overfit (depending on the amount of task specificity in the training signals) the ground-truth labels of downsteam task during the finetuning or final-training stage. This motivates us -- \textit{is there a good middle ground between the two extremes that has a more optimal amount of task specificity?}

In instance-level detection/segmentation tasks, the datasets usually have two types of annotations: localization-based annotations (\textit{e.g.,} boxes, masks) and class labels for those annotations. While semi-supervised methods utilize the information from both kinds of annotations to do pseudo labeling on unlabeled data, we show that such a practice is not optimal for training and would hurt the final performance. Given the pseudo labels that closely mimic ground-truth labels, the model is likely to take an easier optimization path and potentially arrive at a less-favorable local optimum at the period of training. To mitigate the issue, we can disregard either one of the annotation types for the purpose of pseudo labeling. A related example is the conventional practice of pretraining the model on ImageNet \cite{deng2009imagenet} with just image-level class labels. Conversely, the approach of using localization-based annotations (while ignoring class labels) has not been explored previously, which we believe is a promising direction given the nature of detection/segmentation tasks.

In this work, we propose a novel Class-Agnostic Semi-Supervised Learning (CA-SSL) framework for instance-level detection and segmentation. 
CA-SSL is a framework consisting of three cascaded training stages (pseudo labeler training, warmup training, and finetuning) with two detector types (pseudo labeler and target detector). Each stage employs a specific type of training data and a type of detector. Concretely, we first train a pseudo labeler on the labeled dataset with class-agnostic annotations and then use it to generate pseudo labels on unlabeled images. After that, we perform warmup training for the target detector on the unlabeled data with class-agnostic pseudo labels. Finally, we finetune the warmed-up target detector on class-specific ground-truth annotations for a particular instance-level task. Unlike existing state-of-the-art SSL method Soft-Teacher~\cite{xu2021end} that jointly uses labeled and unlabeled data within its single training stage, these two data types are assigned separately to the different stages of CA-SSL, as shown in Fig.~\ref{fig:sketch}. This decoupling strategy provides the warmup model with a good initial solution (learned from unlabeled data) that guides it to maintain a good generalization performance during finetuning.

In our experiments, we carry out evaluations by considering upper-bound model performance on multiple large-scale unlabeled data splits that have different dataset scales, consisting of images from COCO unlabeled~\cite{lin2014microsoft}, Open Images~\cite{kuznetsova2018open} and Places365~\cite{zhou2017places}. Through extensive experiments, we demonstrate that our method can obtain consistent performance gains when using different unlabeled splits ranging from 0.12M to 3.6M images. Owing to the superior effectiveness of our method at consuming large-scale unlabeled data, we are the first successful attempt to improve task performance using an unlabeled dataset with an enormous amount of 3.6M images, in the history of semi-supervised object detection. Moreover, our class-agnostic warmup model trained with warmup training demonstrates excellent transferability to other instance-level detection and segmentation frameworks.
The contributions for this paper are threefold:
%
\begin{compactitem}
    \item We propose a novel class-agnostic semi-supervised learning framework for instance-level detection/segmentation tasks. By leveraging cascaded training stages and class-agnostic pseudo labels, it achieves a more optimal amount of task specificity in the training signals extracted from unlabeled data.

    \item We conduct extensive ablative and comparative experiments on object detection, demonstrating the effectiveness of our method. To the best of our knowledge, we are the first to use unlabeled data at an unprecedented scale of 3.6M for semi-supervised object detection.
    
    \item We demonstrate that our class-agnostic warmup model trained with warmup training can significantly improve the performance on other instance-level detection/segmentation tasks (instance segmentation, keypoint detection, entity segmentation, panoptic segmentation) and frameworks.
\end{compactitem}

\vspace{-2mm}
\section{Related work}
\vspace{-2mm}
\noindent \textbf{Instance-level detection/segmentation.}
Instance-level detection tasks, including object detection~\cite{girshick2014rich,dollar2015fast,DBLP:conf/nips/RenHGS15,DBLP:conf/cvpr/LinDGHHB17,lin2017focal,tian2019fcos,DBLP:conf/nips/DaiLHS16,qi2021multi}, instance segmentation~\cite{he2017mask,DBLP:journals/corr/abs-1803-01534,xie2020polarmask,chen2019tensormask,zhang2020MEInst,qi2020pointins}, and key point detection~\cite{he2017mask,zhou2019objects,sun2019deep,xiao2018simple}, require detecting objects with different instance-level representations such as bounding box, pixelwise mask, and keypoints.
Recently, numerous class-specific panoptic segmentation~\cite{carion2020end,li2021fully,cheng2021per} and class-agnostic entity segmentation~\cite{qi2021open} methods have been developed to perform dense image segmentation by treating all segmentation masks as instances.
Most of instance-level detection research works generally focus on designing more advanced architectures or detection methods that work well on existing labeled datasets. 
Instead, we aim to design a training framework that better utilizes unlabeled images, without modifying the underlying architecture or method. 
This facilitates better understanding of how far current methods can scale with the help of large-scale unlabeled data.

\vspace{1mm}
\noindent \textbf{Semi-supervised detection.} Semi-supervised learning approaches mainly focus on two directions for instance-level detection. 
One is concerned with the consistency-based methods~\cite{jeong2019consistency,tang2021proposal}, which are closely related to self-supervised 
approaches. 
They usually construct a regularization loss by designing some contrastive pretext task~\cite{he2020momentum,chen2021exploring,xie2021detco,henaff2021efficient,wang2021dense,ramanathan2021predet}. Another direction is on pseudo labeling~\cite{radosavovic2018data,zoph2020rethinking,li2020improving,sohn2020simple,wang2018towards,xu2021end}. As the name implies, they leverage a pretrained detector to generate pseudo labels on unlabeled images. The pseudo labels are usually almost identical to the ground-truth labels. Thus, both two types of labels can be used for joint training with similar losses. Our framework is also based on pseudo labeling, but we decouple the semi-supervised pipeline into three cascaded training stages, where each stage employs a specific type of training data (labeled dataset or unlabeled data). Such a design avoids the complicated and careful weighting strategy required to balance unlabeled and labeled data in the joint training scheme of semi-supervised methods~\cite{xu2021end}, where the common issue is to effectively balance the contributions of noisy pseudo labels and ground-truth annotations.


\vspace{1mm}
\noindent \textbf{Class-Agnostic detection and segmentation.}
Class-Agnostic localization~\cite{DBLP:conf/nips/RenHGS15,wang2021unidentified,sharma2021class,kim2021learning,qi2021open} has been widely used in detection and segmentation. One of the most prominent examples is the two-stage detector~\cite{DBLP:conf/nips/RenHGS15}. It mainly has a class-agnostic region proposal network (RPN) and a detection head. RPN predicts numerous high-quality class-agnostic proposals for further classification and localization refinement. Inspired by this design, our method use class-agnostic pseudo labels on unlabeled data for warming up class-specific target detector. As demonstrated by recent works on open-world detection/segmentation~\cite{sharma2021class,kim2021learning,qi2021open}, it can improve the model's generalization on unseen objects. 
Aside from the benefits shown by existing works, in this paper, we present the first evidence that class-agnostic training can significantly bridge the quality gap between pseudo labels and human (ground-truth) labels. This is important because the quality of pseudo labels directly impacts the effectiveness of learning from unlabeled data.

\section{Methodology}
\begin{figure}[t!]
  \centering
  \includegraphics[width=.95\linewidth]{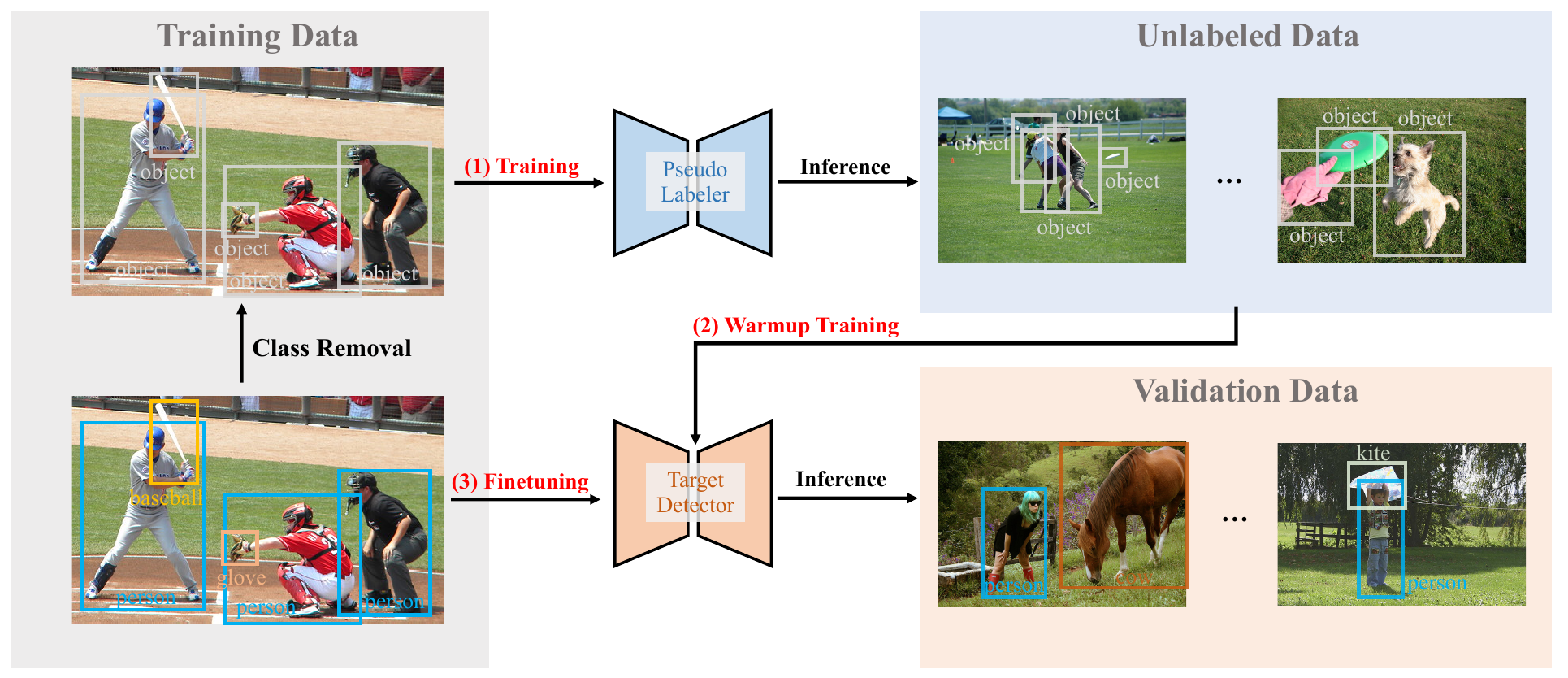}
  \vspace{-2mm}
  \caption{Illustration of 
  CA-SSL framework. The numbered texts indicate the three cascade stages of our framework. It mainly involves two detectors, namely \textit{pseudo labeler} and \textit{target} detector. The bridge between these two detectors is the unlabeled images. We use the pseudo labeler to predict class-agnostic pseudo labels on the unlabeled images, which are then used as training data for warmup training of target detector.}
  \label{fig:pipeline}
  \vspace{-2mm}
\end{figure}

Fig.~\ref{fig:pipeline} provides an overview of the proposed \textbf{C}lass-\textbf{a}gnostic \textbf{S}emi-\textbf{S}upervised \textbf{L}earning (CA-SSL) framework. The framework consists of three stages, including pseudo labeler training, warmup training, and finetuning. In the first stage, we use the labeled data but with only class-agnostic annotations to train a pseudo labeler. This labeling detector then predicts class-agnostic pseudo labels on unlabeled images. In warmup training, these numerous unlabeled images with their pseudo labels are used to train a target detector. This process is akin to pretraining in self-supervised learning with unlabeled data. 
We refer to it as warmup training because only localization-based pseudo labels are used as training data, while class labels are ignored. Finally, we fine-tune the warmed-up target detector on the labeled dataset with class-specific annotations.

In the following sections, we first introduce the entire process of our training framework. After that, we explain in detail our proposal to adopt mask-based annotations at Entity~\cite{qi2021open} level in pseudo labeler and warmup training, as an alternative to the common practice of using box-based object annotations. This annotation adoption is to enable our framework work more effectively beyond instance-level segmentation tasks such as panoptic segmentation. Finally, we describe the base detection framework used by CA-SSL's
training stages.

\vspace{-4mm}
\subsection{Training Stages}
\label{sub::pretraining}

\begin{table}[tp]
\centering
\footnotesize
\setlength{\tabcolsep}{10pt}
\begin{tabular}{c c c c c}
\hline
\cellcolor{lightgray} Train data & \cellcolor{lightgray} Annotation & \cellcolor{lightgray} Num & \cellcolor{lightgray} AP$^{\text{det}}$  & \cellcolor{lightgray} AP$^{\text{det-a}}$ \\ \hline
COCO \texttt{train2017} & Ground-truth  & 118k  & 41.0 & 41.9   \\ \hline
COCO \texttt{unlabeled} & Pseudo labels & 123k  & 35.9 & 40.0   \\ 
\hline
\end{tabular}
\caption{COCO \texttt{val2017} validation results from training on ground-truth and pseudo labels, across class-specific (AP$^{\text{det}}$) and -agnostic (AP$^{\text{det-a}})$ tasks. `Num' is the number of training images. AP$^{\text{det}}$ and AP$^{\text{det-a}}$ indicate the class-specific and -agnostic object detection mAPs respectively.}
\vspace{-10mm}
\label{tab:motivation}
\end{table}


We design our training stages based on the finding that there is only a small quality gap between class-agnostic ground-truth and pseudo labels.
We study the feasibility of class-agnostic detection for pseudo labeling, by contrasting it to the conventional class-specific detection task, in terms of class-agnostic~\cite{qi2021open} and class-specific~\cite{lin2014microsoft} AP metrics respectively. Table~\ref{tab:motivation} shows that the validation performance gap between training on either ground-truth data  (COCO \texttt{train2017}) or pseudo labels (COCO \texttt{unlabeled}), for both class-specific and -agnostic settings\protect\footnotemark on COCO validation set~\cite{lin2014microsoft}. The groundtruth-unlabeled AP gaps of class-specific and class-agnostic models are 5.1\% (41.0-35.9) and 1.9\% (41.9-40.0) respectively. The much smaller gap indicates that the class-agnostic pseudo labels have much a better quality than the class-specific ones, thus enabling the model to closely approach the AP of training on ground-truth labels. In many cases, class-agnostic pseudo labels can eliminate the ambiguities caused by confusing predefined classes such as \textit{cyclist} and \textit{person}, while generalizing well to other kinds of objects unseen , which have been well studied in $\mathcal{D}^t$~\cite{qi2021open}. The class-agnostic model with stronger generalization ability provides a greater variety of proposals which help localize objects better during finetuning. We show that such properties of class-agnostic pseudo labels are more useful for warmup training of the class-specific model, compared to prematurely learning classification-aware features through the class-specific ground-truth annotations of downstream task.

\footnotetext[1]{We use FCOS~\cite{tian2019fcos} with ResNet50 backbone, a widely-used one-stage detector, to explore the performance gap between using class-specific and -agnostic labels. We follow its 36 epoch training setting widely adopted in detectron2~\cite{wu2019detectron2} or mmdetection~\cite{chen2019mmdetection}}

Similar to other semi-supervised learning frameworks~\cite{jeong2019consistency,tang2021proposal,radosavovic2018data,zoph2020rethinking,li2020improving,sohn2020simple,wang2018towards,xu2021end}, our CA-SSL framework is largely based on the conventional object detection training process which we first briefly introduce here. Conventionally, given the input images $I$ and their ground-truth annotations $Y$ from the human-labeled training dataset $\mathcal{D}^t$, the detection model denoted as $h_{*}$ is trained with the composite detection loss:  $\mathcal{L}_\text{det} =  \mathcal{L}_\text{cls} \text{ (classification loss)} + \mathcal{L}_\text{loc} \text{ (localization loss)}$. The detection model $h_{*}$ is learned through a function $\mathcal{H}(*)$ that determines the neural network hypothesis spaces, depending on the task at hand. Next, we provide the details of the three stages of CA-SSL framework.

\vspace{1mm}
\noindent \textbf{Pseudo labeler training.}
The goal of this stage is to train a pseudo labeler on $\mathcal{D}^t$ to generate high-quality class-agnostic pseudo labels $\mathcal{D}^p$ from the unlabeled dataset split $\mathcal{D}^u$. To keep the training and inference consistent, we train the pseudo labeler on the class-agnostic annotations obtained from $\mathcal{D}^t$. We directly remove class information from the annotations in $\mathcal{D}^t$ using the class-agnostic conversion function $\alpha(\cdot)$ and regard each label as a class-free ``object''. To train the pseudo labeler, we use a recent class-specific detection framework (see Sec.~\ref{sec:base_detector}) and replace its multi-class classifier with a binary classifier. Given the labeled dataset $\mathcal{D}^t$, the pseudo labeler $h_\text{L}$ is trained as follows:
\begin{equation}
\begin{split}
h_{\text{L}} = \argmin_{h\in \mathcal{H}(\text{L})} \sum_{\{I_i, Y_i\} \in \mathcal{D}^t} \mathcal{L}_{\text{det}}^a(h(I_i),
\alpha(Y_i)),
\label{eq:L_det}
\end{split}
\end{equation}
\noindent where $\mathcal{L}^a_\text{det}$ representation the class-agnostic version of $\mathcal{L}_\text{det}$ and $\mathcal{H}(\text{L})$ indicates the neural network hypotheses conditioned on the labeling detection task.

Once the training is done, we apply the pseudo labeler to the unlabeled images and then filter the prediction results using merely a single (constant) score threshold $\delta$. Without semantic class labels in the prediction results, class-agnostic pseudo labels avoid the long-tail problem suffered by in class-specific predictions. Some related ablations are in the supplementary file. As a result, there is no need for a complicated strategy that applies class-dynamic score thresholds
as in existing works~\cite{radosavovic2018data,zoph2020rethinking,li2020improving,sohn2020simple,wang2018towards,xu2021end}. Given the pseudo labeler $h_\text{L}$ and score threshold $\delta$, our class-agnostic pseudo labeling process to obtain the pseudo labels $Y^p$ and pseudo-label dataset $\mathcal{D}^p$ is represented by the following:
\begin{equation}
Y^p_i = 
\{y_j \in h_{\text{L}}(I^u_i) \vert \textbf{score} (y_j) > \delta \} \quad \forall I^u_i \in \mathcal{D}^u,
\label{eq:L_det}
\end{equation}
\begin{equation}
\mathcal{D}^p = \{(I^u_i \in \mathcal{D}^u, Y^p_i \in Y^p) | Y^p_i \neq \emptyset\},
\label{eq:L_det}
\end{equation}
\noindent where `$p$' indicates the association with pseudo labels and \textbf{score}($\cdot$) returns the objectness score of any prediction.

\vspace{1mm}
\noindent \textbf{Warmup training.}
We perform warmup training of the target detector only on the pseudo-label dataset $\mathcal{D}^p$. We do not make use of any ground-truth dataset in this stage, which is different from the state-of-the-art semi-supervised approach Soft-Teacher~\cite{xu2021end} that carries out joint training on ground-truth and unlabeled dataset splits. Given that, we do not require a divide-and-conquer strategy to handle different dataset splits, such as applying different loss weights to noisy pseudo labels and clean ground-truth labels~\cite{berthelot2019mixmatch,ren2020not}. To obtain the warmed-up target `T' detection model $h_\text{T}$, we perform warmup training as follows:
\begin{equation}
\small
h_{\text{T}} = \argmin_{h \in \mathcal{H}(\text{T})}\sum_{\{I^p_i, Y^p_i\} \in \mathcal{D}^p}\mathcal{L}^a_{\text{det}}(h(I^p_i), \ceil{Y^p_i}),
\label{eq:L_tdet_p}
\end{equation}
\noindent where $\ceil{\cdot}$ transforms $Y^p$ to binary training targets. Note that 
warmup training is related to the pretraining step of self-supervised learning. The weights of the model are well-initialized for better adaptation to downstream tasks. Compared to self-supervised methods~\cite{chen2020improved,wang2021dense,xie2021detco,henaff2021efficient,ramanathan2021predet}, the unlabeled data with class-agnostic pseudo labels provides relatively more informative and task-specific supervision signals (class-agnostic localization) that significantly facilitate instance-level detection and segmentation tasks.

\vspace{1mm}
\noindent \textbf{Finetuning.}
\label{sub::finetuning}
After obtaining the warmup model $h_\text{T}$ from warmup training, we finetune it for the downstream task using the class-specific ground-truth annotations of $\mathcal{D}^t$. Instance-level detection/segmentation tasks are typically class-specific tasks. Thus, the output channel of target detector's semantic classifier should be adapted to the number of pre-defined classes for the downstream task at hand. There are two ways to initialize multi-class classification layer: (1) random initialization; (2) initialize each output channel with the one from the classification layer of $h_\text{T}$. We empirically find these two strategies produce similar results. The finetuning process to obtain the final `F' downstream-task model $h_\text{F}$ is represented by:
\begin{equation}
\small
\begin{split}
h_{\text{F}} = \argmin_{h \in \mathcal{H}(\text{F}; h_\text{T})}\sum_{\{I_i, Y_i\} \in \mathcal{D}^t}\mathcal{L}_{\text{det}}(h(I_i), Y_i),
\end{split}
\label{eq:L_tdet_f}
\end{equation}

\noindent where $\mathcal{H}(\text{F};h_\text{T})$ indicates that the neural network hypotheses are conditioned on both the task `F' and pretrained model $h_\text{T}$. Our approach of warming up the target detector with only unlabeled data guarantees it less prone to overfitting to the downstream task's images and ground-truth labels, since they have not been exposed to the model in the warmup training stage.

\vspace{-4mm}
\subsection{Switching From Objects to Entities}
\vspace{-1mm}
\label{sec:entity}
\begin{figure}[t!]
  \centering
   \includegraphics[width=0.7\linewidth]{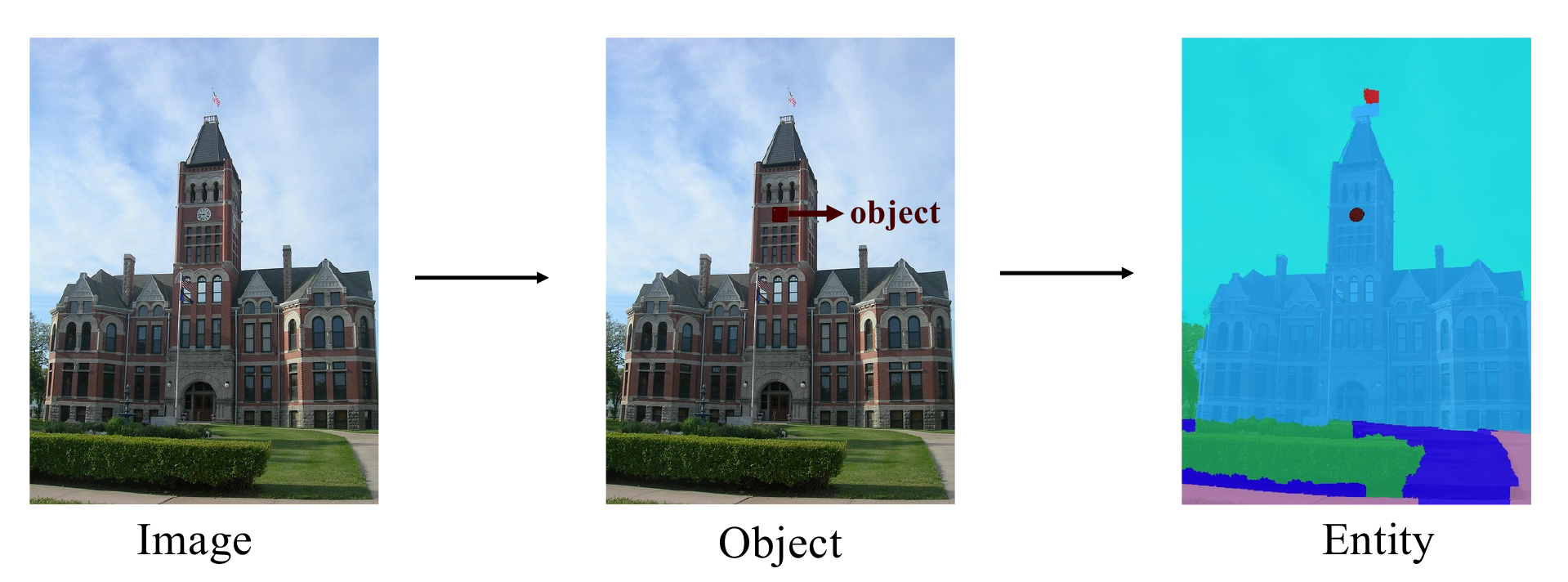}
   \vspace{-4mm}
   \caption{An example showing that switching from \textit{box-based object} annotations to \textit{mask-based Entity} \cite{qi2021open} annotations results in a more substantial set of labels for pseudo labeler and warmup training.}
   \vspace{-5mm}
   \label{fig:entity}
\end{figure}

Warmup training with large-scale unlabeled data is a costly process. 
Thus, it is desirable to design our CA-SSL framework such that the trained class-agnostic model can serve a good range of tasks that expand to detection and also include segmentation. Instance segmentation and panoptic segmentation are two widely-used instance-level segmentation tasks that require more fine-grained visual information for predicting pixelwise masks for each instance.

We draw inspiration from Entity Segmentation \cite{qi2021open} on how to boost the applicability and usefulness of our CA-SSL framework. Instead of just focusing on objects, we propose to perform warmup training for the target detector based on the semantically-coherent and class-agnostic mask regions known as Entities.
Entities include not just object regions but also stuff regions such as \textit{sky} and \textit{road} which come with the panoptic segmentation task in mainstream datasets (\textit{e.g.,} COCO \cite{lin2014microsoft}). With this, even the unlabeled images with little-to-no object regions can still provide substantial pseudo-label training signals through the stuff regions. In Fig.~\ref{fig:entity}, there is only a single object (\textit{clock}), while multiple Entities like \textit{building} and \textit{sky} are present. Furthermore, stuff regions have close relationships with objects, and thus we hypothesize that the training on stuff pseudo labels strongly benefits the downstream task even if it is an object-based task~\cite{mottaghi2014role}.

\vspace{-3mm}
\subsection{Base Detector}\label{sec:base_detector}
We adopt the CondInst~\cite{tian2020conditional}, a widely-used instance segmentation framework, as our base detector for the pseudo labeler and target detector. Different from Mask R-CNN~\cite{he2017mask}, the Condinst is in fully convolutional manner with two parts: a dense one-stage detector FCOS for detection and a segmentation head for mask prediction. The FCOS has a backbone, FPN neck, and a detection head. The detection head has three output branches: the classification, regression, and kernel branch.  The first two branches perform instance-level classification and regression to achieve object detection. Whereas, the kernel branch generates dynamic convolution weights which are used to convolve with high-res feature maps to generate binary instance masks within the segmentation head. 
Such a network architecture keeps the detection and segmentation parts fairly independent, making it easy to transfer its arbitrary parts to other networks and tasks for finetuning, as proved by our ablation study in Table~\ref{Tab:prune} and Table~\ref{Tab:instance}. \textit{E.g.,} we can easily initialize the backbone and FPN neck of Mask R-CNN with our
CondInst target detector without any modifications. Overall, the CondInst base detector is trained with the following:
\begin{equation}
\mathcal{L}^e = \mathcal{L}_{\text{det}}^{\{a,c\}} + \mathcal{L}_{\text{seg}},
\label{eq:L_entity}
\end{equation}
\noindent where $\mathcal{L}_{\text{seg}}$ is usually the dice loss between predicted segmentation mask and ground truth. We choose $\mathcal{L}_{\text{det}}^{\{a,c\}}$ as class-agnostic $\mathcal{L}_{\text{det}}^a$ or class-specific $\mathcal{L}_{\text{det}}^c$ depending on the detector and training stage described in subsection~\ref{sub::pretraining}.

Also, CondInst learns to group the pixels that belong to the same semantic/instance region, and otherwise push them far apart. This can be seen as a form of implicit contrastive learning which focuses on spatial-level representation learning, rather than global vector-based representation. This leads to segregated features which are useful for detection and segmentation tasks.

\vspace{-4mm}
\section{Experiments}\label{sec:exp}
\paragraphx{Datasets.} MS-COCO~\cite{lin2014microsoft} is used as the main evaluation dataset. In addition to MS-COCO \texttt{train2017} and \texttt{val2017} splits,  we curate 3.6 million unlabeled images from COCO \texttt{unlabeled}~\cite{lin2014microsoft}, Places365~\cite{zhou2017places}, and Open Images~\cite{kuznetsova2018open}. To
better demonstrate the data scalability of CA-SSL, we construct four unlabeled data subsets (\texttt{tiny}, \texttt{small}, \texttt{base}, \texttt{large}) with 120K, 660K, 1.74M, and 3.66M images respectively. Unless specified, we report the experimental results from using \texttt{tiny} as the unlabeled subset, which is from COCO \texttt{unlabeled} and has nearly the same scale as \texttt{train2017} as shown in Table~\ref{tab:motivation}.

\paragraphx{Training Setup.} 
For fair comparisons with other methods (some may require less training time), we train all models to reach their respectively upper-bound performances, by increasing the number of training epochs accordingly. Unless specified, we train with 60 and 36 epochs in warmup training and finetuning respectively for CA-SSL to achieve the upper-bound performance. Upper-bound performance evaluation is the preferred way to gauge the true performance of different methods that may require different training costs. We apply either \textit{weak} data augmentation (conventional multi-scale training) or \textit{strong} data augmentation (jittering of scale, brightness, contrast, etc. \cite{xu2021end,chen2021scale}) to the unlabeled images before generating pseudo labels.

\paragraphx{Implementation.} 
We adopt the Condinst~\cite{tian2020conditional} framework with Swin Transformer Tiny (T) backbone~\cite{liu2021swin} as our base model. Please refer to our supplementary file for the hyper-parameter settings in the warmup training and finetuning.
\vspace{-4mm}
\subsection{Experimental Results}
\begin{table}[t!]
    \centering
    \footnotesize
    \setlength{\tabcolsep}{3pt}
    \begin{tabular}{c|cccc|ccc}
    \cellcolor{lightgray} warmup & \multicolumn{4}{c}{\cellcolor{lightgray} warmed-up model} & \multicolumn{3}{c}{\cellcolor{lightgray} finetuned model} \\ \cline{2-8}
    \cellcolor{lightgray} training data & \cellcolor{lightgray} AP$^{\text{det-a}}$ & \cellcolor{lightgray} AP$^{\text{seg-a}}$ & \cellcolor{lightgray} AP$^{\text{det}}$ & \cellcolor{lightgray} AP$^\text{e}$ & \cellcolor{lightgray} AP$^\text{det}$  & \cellcolor{lightgray} AP$^\text{seg}$ & \cellcolor{lightgray} PQ \\ 
    ImageNet & -  & -  & - & - & 46.8  & 43.0 & 41.4\\ 
    Objects(G,A) & 41.9 & 35.3 & - & - & 46.9 & - & - \\ 
    Objects(P,S) & - & - & 35.9 & - & 47.1 & - & - \\ 
    Objects(P,A) & 40.0 & 33.4 & -  & - & 48.1  & 45.0 & 42.5 \\ 
    Entities(P,A) & - & - & - & 40.2  & \textbf{48.2} & \textbf{45.2} & \textbf{43.7} \\ \hline
    \end{tabular}
    \caption{Effect of warmup training data choice. AP$^{\text{det}}$ and AP$^{\text{seg}}$ indicate the APs for class-specific object detection and instance segmentation tasks. Their class-agnostic counterparts have names appended with `-a'. AP$^{\text{e}}$ and PQ are the evaluation metrics of class-agnostic Entity detection \cite{qi2021open} and panoptic segmentation. `G' and `P' indicate whether ground-truth (COCO \texttt{train2017} with 118K images) or pseudo labels (COCO unlabeled data with 123K images) are used. They are different datasets but with comparable numbers of images.   `A' and `S' indicate whether class-agnostic or class-specific labels are used. Note that all the results here are obtained via COCO \texttt{val2017}.}
    \vspace{-4mm}
    \label{Tab:exp_trans_entity}
\end{table}

\begin{table}[t!]
\centering
\footnotesize
\setlength{\tabcolsep}{8pt}
\begin{tabular}{c|ccc|ccc}
\cellcolor{lightgray} epochs & \cellcolor{lightgray} AP$^\text{e}$  & \cellcolor{lightgray} AP$^\text{e}_{50}$ & \cellcolor{lightgray} AP$^\text{e}_{75}$ & \cellcolor{lightgray} AP$^\text{e}_{\text{s}}$ & \cellcolor{lightgray} AP$^\text{e}_{\text{m}}$ & \cellcolor{lightgray} AP$^\text{e}_{\text{l}}$\\ 
12 (1$\times$) & 38.4 & 59.6 & 40.5 & 14.5 & 35.0 & 52.6   \\ 
24 (2$\times$) & 39.6 & 60.5 & 41.9 & 15.0 & 36.3 & 54.1   \\ 
36 (3$\times$) & 39.9 & 60.8 & 42.4 & 15.1 & 36.9 & 54.6   \\ 
48 (4$\times$) & 40.2 & 60.8 & 42.7 & 15.4 & 37.1 & 54.8   \\ 
60 (5$\times$) & \textbf{40.2} & \textbf{60.9} & \textbf{42.8} & \textbf{15.4} & \textbf{37.0} & \textbf{54.9}   \\ \hline
\end{tabular}
\caption{Performances of pretrained model with different number of warmup training epochs. The AP$^{\text{e}}_{50}$ and AP$^{\text{e}}_{75}$ are APs of the Entity detection task based on 0.5 and 0.75 IoU thresholds. The AP$^{\text{e}}_{\text{s}}$, AP$^{\text{e}}_{\text{m}}$ and AP$^{\text{e}}_{\text{l}}$ indicate the mAP performance on small, middle, and large entities.}
\label{tab::upper_pretrain}
\vspace{-9mm}
\end{table}

\paragraphx{Warmup Training Data.}
Table~\ref{Tab:exp_trans_entity} shows the impact of the choice of warmup training data on the downstream task performances after the finetuning stage. From the first two rows, we observe that ImageNet supervised-pretrained model and the warmup model trained on class-agnostic ground-truth labels suffer from the worst downstream performances. Whereas, in the last three rows, the models that leverage pseudo labels `P' for training consistently achieve stronger downstream performances. It can also be clearly seen that using class-agnostic labels `A' during warmup training is better than using class-specific labels `S', resulting in a downstream AP$^{\text{det}}$ gap of 1.0\% (48.1-47.1). Class-Agnostic warmup training helps to prevent the model from overfitting to the ground-truth labels during finetuning. In particular, Entity-based warmup training data provides the best overall downstream task performance, while strongly raising the downstream PQ performance by 1.2\% (43.7-42.5) due to the inclusion of both object and stuff elements. Note that the last two rows of Table~\ref{Tab:exp_trans_entity} use identical COCO \texttt{unlabeled} images but with different annotation types: box-based objects and mask-based entities~\cite{qi2021open} (objects $\&$ stuffs), as illustrated in Sec.~\ref{sec:entity}.

\paragraphx{Upper-Bound Performance.} We investigate the number of training epochs required to achieve the upper-bound performance in both the warmup training and finetuning stages. Table~\ref{tab::upper_pretrain} shows the class-agnostic performance under different numbers of warmup training epochs. When increasing the number of epochs from 12 to 48, the performance of trained model improves from 38.4 to 40.2 in terms of class-agnostic AP$^\text{e}$. The improvement saturates after 40 epochs. Table~\ref{tab::upper_finetune} shows the the downstream class-specific object detection performance under different combinations of warmup training and finetuning epochs. With 60 epochs (5$\times$) in the warmup training and 36 epochs (3$\times$) in finetuning stages, we obtain the best performance of 48.2 AP$^{\text{det}}$. Therefore, we adopt this particular setting in the rest of the experiments.

\begin{table}[t!]
\centering
\footnotesize
\setlength{\tabcolsep}{8pt}
\begin{tabular}{cccccc}
\cellcolor{lightgray} \diagbox[height=5mm, width=18mm]{F}{W}& \cellcolor{lightgray}1$\times$ & \cellcolor{lightgray} 2$\times$ & \cellcolor{lightgray} 3$\times$ &
\cellcolor{lightgray} 4$\times$ & 
\cellcolor{lightgray} 5$\times$ \\ 
1$\times$ & 45.3 & 46.3 & 46.4 & 46.8 & 47.1  \\ 
2$\times$ & 47.0 & 47.4 & 47.7 & 47.3 & 47.7 \\ 
3$\times$ & 47.2 & 47.7 & 47.6 & 47.9 & \textbf{48.2}   \\ 
4$\times$ & 47.3 & 47.6 & 47.7 & 47.9 & 48.0   \\ 
5$\times$ & 47.5 & 47.7 & 47.9 & 47.9 & 48.0  \\ \hline
\end{tabular}
\caption{Exploration to the best training epoch combination in the warmup training and finetuning stages on downstream object detection performance AP$^{\text{det}}$. E$\times$ indicates the E$\times$12 training epochs. \textit{E.g.}, 1$\times$ and 5$\times$ represent 12 and 60 training epochs, respectively. `F' and `W' represent the \textbf{F}inetuning and \textbf{W}armup training stages.}
\vspace{-4mm}
\label{tab::upper_finetune}
\end{table}

\footnotetext{ResNet-101, Swin-Tiny, Swin-Small, Swin-Base, and Swin-Large.}
\paragraphx{Pseudo labeler.}
We investigate the impact of pseudo labeler in the first training stage. Table~\ref{tab::label_model} shows how the performance of pseudo labeler affects the performance of the target detector (after warmup training is performed) on the class-agnostic entity detection task. The results here conclude that a stronger pseudo labeler consistently leads to a stronger target detector in warmup training. 
\begin{table}[t!]
\centering
\footnotesize
\setlength{\tabcolsep}{8pt}
\begin{tabular}{cccccc}
\cellcolor{lightgray} pseudo labeler (AP$^\text{e}$) & 
\cellcolor{lightgray} 38.0 & 
\cellcolor{lightgray} 39.1 & 
\cellcolor{lightgray} 40.6 & 
\cellcolor{lightgray} 41.4 & 
\cellcolor{lightgray} \textbf{42.1} \\ 
warmed-up target detector (AP$^\text{e}$) &36.1 & 37.3 & 38.8 & 39.6 & \textbf{40.2}  \\ \hline
\end{tabular}
\caption{Relationship between the performance of pseudo labeler and the performance of the warmed-up target detector right after warmup training. The different columns correspond to the various backbones\protect\footnotemark used by the pseudo labeler. The warmup target detector is always based on \texttt{Tiny} backbone.}
\vspace{-9mm}
\label{tab::label_model}
\end{table}

\paragraphx{Data Augmentation.} Table~\ref{tab::aug} shows the impact of data augmentation scheme adopted in the warmup training and finetuning stages, on the intermediate and final detection performances. Using strong data augmentation independently for any of the two stages brings some performance improvement, while combining the two provides the largest AP$^\text{det}$ improvement of $0.9\%$ (49.1-48.2). 

\begin{table}[t!]
\centering
\footnotesize
\setlength{\tabcolsep}{7pt}
\begin{tabular}{ccc}
\cellcolor{lightgray} \diagbox[height=5mm, width=18mm]{F}{W} & 
\cellcolor{lightgray} weak & 
\cellcolor{lightgray} strong \\ 
weak & (40.2, 48.2) & (\textbf{40.6}, 48.6)  \\ 
strong & (40.2, 48.6) & (\textbf{40.6}, \textbf{49.1}) \\ \hline
\end{tabular}
\caption{Ablation study on data augmentation. 
`W' and `F' represent the warmup training and finetuning stages, while `weak' and `strong' refer to weak and strong data augmentation. (A, B) indicate the class-agnostic AP$^\text{e}$ and class-specific AP$^{\text{det}}$ in warmup training and finetuning.}
\vspace{-4mm}
\label{tab::aug}
\end{table}

\paragraphx{Scale of Unlabeled Dataset.} Table~\ref{tab:unlabeled_nums} shows how the performance varies by training the model on different unlabeled dataset splits. With the increase of unlabeled dataset scale, the performances of the models from the warmup training and finetuning stages improve consistently. 
We also notice that the class-agnostic performance of the warmup model correlates well with the performance of the finetuning model. 
%
%
This is expected as the substantial task similarity between those two training stages.

\begin{table}[t!]
\centering
\footnotesize
\begin{tabular}{c|c|c|c}
\cellcolor{lightgray} Setting & 
\cellcolor{lightgray} Num & 
\cellcolor{lightgray} warmup model (AP$^{\text{e}}$) & 
\cellcolor{lightgray} fine-tuned model (AP$^{\text{det}}$)    \\ 
Tiny    & 123K   & 40.2 & 48.2                     \\ 
Small   & 660k   & 40.7 & 48.8                     \\ 
Base    & 1.74M  
& 41.4 & 49.7                     \\ 
Large   & 3.66M  
& \textbf{42.0} & \textbf{50.6}   \\ \hline
\end{tabular}
\caption{Ablation study on unlabeled splits with different dataset scales.}
\label{tab:unlabeled_nums}
\end{table}

\begin{table}[t!]
\scriptsize
\begin{center}
\begin{tabular}{c|c|c|c|c}
\cellcolor{lightgray} Type & 
\cellcolor{lightgray} Method & 
\cellcolor{lightgray} Unlabeled Images &
\cellcolor{lightgray} Model  &  
\cellcolor{lightgray}AP$^{\text{det}}$ \\
\multirow{5}*{Self-supervised} & MoCo-v2~\cite{chen2020improved} & 1.28M & R-50-FPN &
$39.7\xrightarrow{+0.1}39.8$\\
& DenseCL~\cite{wang2021dense} & 1.28M & R-50-FPN & $39.7\xrightarrow{+0.6}40.3$\\
& DetCo~\cite{xie2021detco} & 1.28M & R-50-MaskRCNN & $38.9\xrightarrow{+1.2}40.1$\\ 
& DetCon~\cite{henaff2021efficient} & 1.28M & R-50-FPN & $41.6\xrightarrow{+1.8}42.7$\\
& PreDet~\cite{ramanathan2021predet} & 50.00M & R-50-MaskRCNN & 
$44.9\xrightarrow{+2.2}47.1$\\ \hline \hline
\multirow{10}*{Semi-supervised} & Proposal learning~\cite{tang2021proposal} & 0.12M & R-50-FPN & $37.4\xrightarrow{\text{+1.0}}38.4$\\
& STAC~\cite{sohn2020simple} & 0.12M & R-50-FPN & $39.5\xrightarrow{\text{-0.3}}39.2$ \\
& Self-training~\cite{zoph2020rethinking} & 2.90M & R-50-FPN (SimCLR) &  $41.1\xrightarrow{\text{+0.8}}41.9$ \\ \cline{2-5}

& \multirow{2}{*}{Soft Teacher~\cite{xu2021end}} & 1.74M$^\dagger$ & Swin-T-FCOS & $46.8\xrightarrow{+2.5}49.3$\\
& & 3.66M$^\ddagger$ & Swin-L-HTC++ & $58.2\xrightarrow{+1.7}59.9$\\ \cline{2-5}
& \multirow{3}{*}{CA-SSL (ours)} & 1.74M$^\dagger$ & \multirow{2}{*}{Swin-T-FCOS} & $46.8\xrightarrow{+4.0}50.8$\\ \cline{3-3}\cline{5-5}
& & \multirow{2}{*}{3.66M$^\ddagger$} &  & $46.8\xrightarrow{+4.7}51.5$ \\ \cline{4-5}
& & & Swin-L-HTC++  & $58.2\xrightarrow{+2.7}60.9$ \\ \hline
\end{tabular}
\end{center}
\caption{Comparison with the state-of-the-arts under the setting of \texttt{train2017} set with 118k images. The `unlabeled images' means the number of unlabeled images we use. In the column of `AP$^{\text{det}}$', the left part is the baseline performance of using only \texttt{train2017}, and the right part is the performance of using both \texttt{train2017} and unlabeled dataset. $\rightarrow$ indicates the performance gain. The symbols $^\S$(ImageNet), $^*$(COCO), $^\dagger$\ $^\ddagger$(our curated sets described in Sec.~\ref{sec:exp}) refer to the same respective sets of unlabeled images.}
\vspace{-8mm}
\label{tab::sota}
\end{table}

\paragraphx{State-of-the-art Comparison.} 
In Table~\ref{tab::sota}, we compare the performance of CA-SSL on the downstream object detection task with those of state-of-the-art methods in strong data augmentation. 
Our method achieves the most significant performance gains even though our model is based on an already strong baseline with the powerful Swin-Tiny backbone. With 1.74M unlabeled images, we obtain 50.8 AP$^{\text{det}}$ with 4.0$\%$ improvement over the ImageNet-pretrained baseline. Furthermore, by increasing the number of unlabeled images to 3.66M, we observe an even bigger performance gain of 4.7$\%$ that leads to 51.5 AP$^{\text{det}}$. %
Since there is no obvious sign of performance saturation, we believe that using a super-scale unlabeled dataset (larger than our 3.66M one) can potentially improve model performance significantly. 

Moreover, using a stronger detector Swin-Large-HTC++\cite{chen2019hybrid} with CA-SSL still provides a meaningful 2.7\% improvement (58.2 $\rightarrow$ 60.9 AP$^{\text{det}}$), suggesting that CA-SSL is compatible with advanced detectors. With the same detector and similar unlabeled images, Soft Teacher~\cite{xu2021end} merely achieves a 2.5 $\%$ (Swin-Tiny-FCOS) and 1.7 $\%$ (Swin-Large-HTC++) gain, despite its relatively significant 3.6$\%$ gain with R-50-FPN backbone and 120K unlabeled images. 
The reasons are twofold. 
Unlike CA-SSL that decouples the usage of different source data into three stages, Soft Teacher performs joint training and that requires it to be particular about the scale of unlabeled data, in order to achieve a good balance between the contributions from labeled and unlabeled data. Using an unlabeled dataset larger than expected can spoil such a balance. On the other hand, Self-training~\cite{zoph2020rethinking} performs much worse than ours, even with a large amount of 2.9M unlabeled images. This is due to the premature use of class/downstream-specific labels in the pseudo-labeling stage, potentially causing the model overfit easily to ground-truth labels during finetuning. Whereas, CA-SSL mitigates such a problem by not allowing the model to train on labeled images and class-specific labels during warmup training. 

Note that the performance improvements of our proposed semi-supervised learning do not come from Transformer architecture. With R-50-FPN detector and similar 0.12M unlabeled images, CA-SSL improves the performance at a much larger margin than other semi-supervised methods.

\paragraphx{Initialization Strategy.}
\begin{table}[t!]
\centering
\footnotesize
\begin{tabular}{ccccccc}
  \hline
\cellcolor{lightgray}  backbone & 
\cellcolor{lightgray} neck & 
\cellcolor{lightgray} head & 
\cellcolor{lightgray} classifier & \cellcolor{lightgray} AP$^{\text{det}}$ & \cellcolor{lightgray} AP$_{50}^{\text{det}}$ & \cellcolor{lightgray} AP$_{75}^{\text{det}}$ \\ \hline
  $\circ$ & $\circ$ & $\circ$ & $\circ$ & 46.8 & 66.2 & 50.8\\ \hline
  \checkmark & $\circ$ & $\circ$ & $\circ$ & 50.6 & 69.2 & 55.2\\ \hline
  \checkmark & \checkmark & $\circ$ & $\circ$  & 50.8 & 69.4 & 55.5\\ \hline
  \checkmark & \checkmark  & \checkmark & $\circ$ & \textbf{51.5} & 69.5 & 55.3\\ \hline
  \checkmark & \checkmark & \checkmark & \checkmark & \textbf{51.5} & \textbf{69.6} & \textbf{55.4} \\ \hline
\end{tabular}
\caption{Ablation study on the initialization strategy. The pretrained model is divided into four parts here. $\circ$ indicates the particular part's pretrained weights are not being used, while $\checkmark$ indicates otherwise.} 
\vspace{-6mm}
\label{Tab:prune}
\end{table}
We study the effects of including/excluding the multiple parts (backbone, neck, head, and classifier) of the pretrained model from the warmup training stage for initializing the finetuning model. 
Table~\ref{Tab:prune} shows that the main source of improvement (+3.8$\%$) comes from backbone initialization, while the other parts make smaller improvements. 
This ablation study motivates us to transfer our pretrained model to other instance-level detection/segmentation tasks that may not use the downstream frameworks as ours.

\begin{table}[t!]
    \centering
    \scriptsize
    \setlength{\tabcolsep}{5pt}
    \centering
    \label{Tab:cc}
    \begin{tabular}{c|c|c|c|L{0.2cm}L{0.8cm}}
        \hline
\cellcolor{lightgray} Task & 
\cellcolor{lightgray} Type & 
\cellcolor{lightgray} Framework & 
\cellcolor{lightgray} Pretrained w/ &  \multicolumn{2}{c}{\cellcolor{lightgray} Task Perf.} \\ \hline
        \multirow{6}*{DET}
        & \multirow{2}*{E} & \multirow{2}*{FCOS~\cite{tian2019fcos}} & ImageNet & 46.8 & \\
        & & & Ours & 51.5 & \textcolor[RGB]{34,139,34}{(+4.7)} \\ \cline{2-6}
        & \multirow{4}*{P} & \multirow{2}*{RetinaNet~\cite{lin2017focal}} & ImageNet & 42.8  \\ 
        & & & Ours & 47.9 & \textcolor[RGB]{34,139,34}{(+5.1)}\\ \cline{3-6}
        & &  \multirow{2}*{FPN~\cite{DBLP:conf/cvpr/LinDGHHB17}} & ImageNet & 44.0  \\ 
        & & & Ours & 46.5 & \textcolor[RGB]{34,139,34}{(+2.5)} \\ \hline
        
        \multirow{4}*{INS} 
        & \multirow{2}*{E} & \multirow{2}*{CondInst~\cite{tian2020conditional}} & ImageNet & 41.9  \\ 
        & & & Ours & 44.9 & \textcolor[RGB]{34,139,34}{(+3.0)} \\ \cline{2-6}
        & \multirow{2}*{P} & \multirow{2}*{Mask R-CNN~\cite{he2017mask}} & ImageNet & 42.8  \\ 
        & & & Ours & 45.5 & \textcolor[RGB]{34,139,34}{(+2.7)}\\ \hline

        \multirow{2}*{Entity}
        & \multirow{2}*{E} & \multirow{2}*{CondInst~\cite{tian2020conditional}} & ImageNet & 35.1  \\ 
        & & & Ours & 38.2 & \textcolor[RGB]{34,139,34}{(+3.1)}\\ \hline

        \multirow{2}*{Point}
        & \multirow{2}*{P} & \multirow{2}*{Mask R-CNN~\cite{he2017mask}} & ImageNet & 66.8  \\ 
        & & & Ours & 67.8 & \textcolor[RGB]{34,139,34}{(+1.0)} \\ \hline

        \multirow{2}*{Panop}
        & \multirow{2}*{P} & \multirow{2}*{PanopticFPN~\cite{kirillov2019panoptic}} & ImageNet & 39.5   \\ 
        & & & Ours &41.5 &\textcolor[RGB]{34,139,34}{(+2.0)}\\ \hline
        
    \end{tabular}
    \caption{Evaluation on representative instance-level detection/segmentation tasks with Swin-Tiny backbone. ``E'' and ``P'' indicate initializing the framework with \textit{entire} or \textit{part(s) of} our pretrained model. ``ImageNet" and ``Ours'' refer to ImageNet and our  weights trained in the warmup stage. ``Task Perf'' refers to the task-specific evaluation metrics, including AP for object detection, instance segmentation and key-point detetcion, AP$^{\text{e}}$ for entity segmentation, and PQ for panoptic segmentation.}
    \vspace{-6mm}
    \label{Tab:instance}
\end{table}

\paragraphx{Generalization to Other Frameworks/Tasks.} Table~\ref{Tab:instance} shows the strong performance improvements resulted from initializing the downstream models with a single model $h_\text{T}$ pretrained with CA-SSL, on various instance-level detection and segmentation tasks. Some frameworks like Mask R-CNN have 
heads that are incompatible with those of our Condinst base detector. Using just our FPN backbone pretrained weights to initialize Mask R-CNN, we achieve 2.7 AP$^{\text{seg}}$ and 1.0 AP$^{\text{point}}$ improvements on instance segmentation and keypoint detection. This demonstrates the strong generalization ability of our semi-supervised learning method.

\vspace{-4mm}
\section{Conclusion}
In this work, we propose a class-agnostic semi-supervised learning framework to improve instance-level detection/segmentation performance with unlabeled data. 
To extract the training signals with a more optimal amount of task specificity, the framework adopts class-agnostic pseudo labels and includes three cascaded training stages, where each stage uses a specific type of data. By performing warmup training on a large amount of class-agnostic pseudo labels on unlabeled data, the class-agnostic model has strong generalization ability and is equipped with the right amount task-specific knowledge. When finetuned on different downstream tasks, the model can better avoid overfitting to the ground-truth labels and thus can achieve better downstream performance. Our extensive experiments show the effectiveness of our framework on object detection with different unlabeled splits. Moreover, the pretrained class-agnostic model demonstrates excellent transferability to other instance-level detection frameworks and tasks.

\clearpage
\bibliographystyle{splncs04}
\bibliography{egbib}

\end{document}